# Capitalization and Punctuation Restoration: a Survey

Vasile Păiș , Dan Tufiș

Research Institute for Artificial Intelligence "Mihai Drăgănescu", Romanian Academy
CASA ACADEMIEI, 13 "Calea 13 Septembrie", Bucharest 050711, ROMANIA
{vasile, tufis}@racai.ro

**ABSTRACT**

Ensuring proper punctuation and letter casing is a key pre-processing step towards applying complex natural language processing algorithms. This is especially significant for textual sources where punctuation and casing are missing, such as the raw output of automatic speech recognition systems. Additionally, short text messages and micro-blogging platforms offer unreliable and often wrong punctuation and casing. This survey offers an overview of both historical and state-of-the-art techniques for restoring punctuation and correcting word casing. Furthermore, current challenges and research directions are highlighted.

## 1 Introduction

### 1.1 Motivation

Text understanding by humans and artificial intelligence (AI) programs often require the proper use of capital letters and punctuation marks. For simple sentences, composed of few words, it may be possible for both humans and AI to read and process even if capital letters usage or punctuation is not present or is inconsistent. For example, in the case of voice-based commands, processing is usually performed only on recognized lower-case words. However, if the body of text to be analyzed increases, such as whole paragraphs or pages, then even for humans it is a challenging task to quickly comprehend its meaning. This was studied by Jones et al. (2003), who analyzed the impact of capitalization and punctuation on the readability of speech-to-text transcripts.

Early work considered punctuation only as cues from the reader perspective to the possible prosodic and pause characteristics of the text (Markwardt, 1942). Nunberg (1990) argues for a much greater role associated with punctuation. Furthermore, punctuation marks are classified as delimiting, separating and disambiguating. Some marks, like the comma, may belong to multiple categories since they are able to perform multiple roles. Jones (1994) proves that "for longer sentences of real language a grammar which makes use of punctuation massively outperforms an otherwise similar grammar that ignores it". Based on this and other similar findings, current state of the art language models consider punctuation as part of their vocabulary. This includes recent models such as BERT (Devlin et al, 2018), ELMo (Peters et al, 2018), OpenAI's GPT-2 (Radford et al, 2019) and we can make an educated guess about GPT-3 (Brown et al, 2020) (the model itself as well as the source code are not publicly available).

Natural language processing (NLP) algorithms such as named entity recognition (NER), part-of-speech (POS) identification, dependency parsing, machine translation (MT) make use of capital letters as features regarding the currently processed word while punctuation is used as features for adjacent words. For example, the Stanford Named Entity Recognizer (Finkel et al., 2005) considers features based on word shape. This means constructing a word representation based on the type of characters appearing in the word. Multiple algorithms for word shape representation were proposed, but the general idea is to encode an uppercase letter with a specific character, say "X", a lowercase letter with "x" and a digit with "d". In this case a word like "McDonald" would become "XxXxxxx". Any such algorithms or variants are possible only if the words are properly represented in terms of upper and lowercase letters. Additionally, the work of Spitkovsky et al. (2011) studies the impact of punctuation on unsupervised dependency parsing.

Special consideration must be given to Automatic Speech Recognition (ASR) systems. Primary output of such systems usually consists of raw text, using the same casing (either lowercase or uppercase) and without any punctuation. In such situations, before applying further NLP algorithms, additional preprocessing is required in order to restore proper letter case and punctuation. These are sometimes referred to as "rich transcriptions". One of the first initiatives concerning automatic rich transcriptions of spoken language started in 2002 in the context of DARPA Effective, Affordable Reusable Speech-to-text (EARS)[1] programme which supported the goal of significantly advancing the state-of-the-art in the hardest speech recognition challenges, including rich transcriptions of language. To this end, NIST released a series of rich transcription evaluation data sets[2], to aid in the evaluation of such systems.

Even though a large volume of data requiring capitalization and punctuation restoration comes from ASR systems, other sources must be considered as well. Miller et al. (2000) identify other noisy sources in the form of text obtained via optical character recognition (OCR) or in some newspaper articles. In these cases, not all the text is affected by lack of proper letter or punctuation but parts of it. In the case of OCR, it is possible that some punctuation is not recognized, while in the case of certain articles it is possible for the first sentence or paragraph to be written using only capital letters. Additionally, in the case of short text messages (SMS), chats, tweets or other micro-blogging activities it is also possible for people to ignore proper casing and punctuation (Baldwin et al., 2013), (Nebhi et al., 2015).

Capitalization plays an important role in systems producing uncased output. For example, in Sadat et al. (2005) is presented the PORTAGE machine translation system. The uncased output of the system can benefit from capitalization restoration as described in Agbago et al. (2005).

In the construction of human-computer interfaces using natural language, sometimes referred to as "chatbots", one of the difficulties encountered is represented by inconsistent use of punctuation and capitalization by the user (Coniam, 2008). In this context, many approaches try to hide the issue by removing all punctuation and capitalization from both training and runtime data (Shawar and Atwell, 2007). Furthermore, Coniam (2014) analyzed also the output text of chatbots from the perspective of a human using these programs to learn English as a second language. He was able to identify issues with capitalization and punctuation even in the produced text. Nevertheless, he argues that for short sentences produced by the chatbots, "changes to English through the increasing adoption of texting makes it debatable whether these issues can be considered important nowadays".

## 1.2 Task description

Capitalization, also known as "truecasing", focuses on identifying the correct word form, making a distinction between the following four classes: all letters lowercase (regular words inside sentences), all letters uppercase (usually the case for certain abbreviations), only the first letter uppercase (starting words of sentences, most proper nouns) and mixed case comprised of several uppercase letters and several lowercase letters (this is the case for certain proper nouns, like "McDonald"). Sometimes an additional fifth class may be used: "no case", considering strings for which there is no upper/lower case distinction, such as numbers or symbols.

Punctuation restoration is the task of inserting appropriate punctuation marks in the appropriate positions, considering an input text without any punctuation. Usually the following punctuation marks are considered: comma, period (or full stop), exclamation mark, question mark, colon, semicolon and quotation marks. However, since commas and periods occur a lot more than the others, lots of studies focus only on these ones. It must be noted that the first letter uppercase due to the word being the first in a sentence does imply a sentence segmentation mechanism, which is usually making use of punctuation restoration. Therefore, the two tasks are intertwined and even though most studies focus on only one of the tasks, there are studies focusing on both using integrated approaches.

In certain studies, only the sentence segmentation sub-task was considered (Stolcke and Shriberg, 1997). Furthermore, Matusov et al. (2006) suggest that it is much easier to predict segment boundaries first and then recover punctuation on the segments instead of the initial unsegmented text. Another important observation is in the case of machine translation software where the translation may be performed on the text segments and punctuation is recovered in the target language where conventions for punctuation may differ compared to the source language.

---

[1] http://mi.eng.cam.ac.uk/research/projects/EARS/ears_summary.html

[2] https://www.nist.gov/itl/iad/mig/rich-transcription-evaluation

## 1.3 Outline

The present study will focus on general text processing for the purpose of case and punctuation restoration. Nevertheless, since there are methods specific to ASR, which make use of audio information, such as pauses in speech, these will be presented as well.

Section 2 presents methods for capitalization and punctuation restoration using only lexical features, starting from rule-based approaches, n-gram language models, discrimination decision, hidden event language models, boosting, conditional random fields and neural networks. Section 3 presents methods that incorporate features specific to audio files. Section 4 considers evaluation resources and methods used for the considered tasks. Finally, section 5 presents the conclusions associated with this work.

# 2 Methods using lexical features

## 2.1 Rule-based approaches

One of the early implementations for a rule-based approach to automatic capitalization is included in the popular word processor Microsoft Word. In this case, a sentence capitalization state machine is used to determine the start of a new sentence, hence indicating the starting word should be capitalized. The details for this approach are described in U.S. patent 5761689 (Rayson et al., 1998). In this case, an analysis is performed on the characters of the current word, defined as regular text characters (letters and digits) and any adjacent punctuation. If during the analysis is identified a punctuation mark indicating end of a sentence, the state machine changes its state accordingly thus being able to suggest a capital letter should follow. Furthermore, if a word is determined as not being the first in a sentence and it is formed of only lowercase letters, then the most frequent capitalization form of the word is assigned, based on dictionary lookup.

Apart from capitalization at the beginning of sentences, inside the sentence capitalization may be required especially in the case of proper names. A possible approach for word capitalization inside sentences is based on building a list of unambiguous capitalized words from the document and suggest based on this list capitalization where it is not used. This approach is described in (Mikheev, 1999).

In (Kim and Woodland, 2002) is presented a system making use of a rule based named entity recognizer. The assumption is that most capitalized words are either first words in sentences or part of a named entity. Most of current named entity recognizers are usually used later in the processing pipeline, requiring additional features such as part-of-speech tagging, like Stanford NER (Finkel et al., 2005) or NCRF++ (Yang and Zhang, 2018), or at least sentence boundaries, like the system of Baevski et al. (2019). However, systems such as (Kim and Woodland, 2000) are constructed to work only on raw text without capitalization or punctuation. This particular system uses Brill inference rules (Brill, 1993).

Even though the purpose of this study is to explore capitalization and punctuation restoration techniques and not rule-based named entity recognition systems, since these last ones proved to be useful, some other named entities systems that ignore punctuation must also be considered, such as (Appelt et al., 1995), (Petasis et al., 2001), (Todorovic et al., 2008), (Elsayed and Elghazaly, 2015) and (Tomita et al., 2005). This last one also makes use of speech features, thus being useful if capitalization and punctuation restoration is applied on the results of ASR.

Rule based systems are known to be hard to maintain as they require constant addition of new rules. However, approaches like the one described in (Petasis et al., 2001) allow for a basic set of entities and rules to be expanded by using large corpora in which it is possible to identify new rules that can be added to the rules database. This is usually known as "bootstrapping", while starting entities and rules are known as "seeds". In (Ehara et al., 2013), bootstrapping is defined as a method for harvesting "instances" similar to given "seeds" by recursively harvesting "instances" and "patterns" by turns over corpora using the distributional hypothesis (Harris, 1954). In the same paper it is demonstrated that the performance of bootstrapping algorithms depends on the selection of seeds.

## 2.2 N-gram based language models

A language model (LM) is a model assigning probabilities to sequences of words (Jurafsky and Martin, 2008). An n-gram is a sequence of "n" words. For certain values of "n" these are better known under specific names such as: "unigram" for n=1, "bigram" for n=2, "trigram" for n=3. Such a LM usually implies computing the probability of a word $W_k$ given a context window of n previous words $W_{k-1}$, $W_{k-2}$...$W_{k-n}$. This probability can be expressed by: $P(W_k|W_{k-1},W_{k-2},...,W_{k-n})$.

In (Beeferman et al.,1998) is proposed a LM based on trigrams, extended from a baseline model constructed based only on words, to account also for punctuation. The paper focuses on restoring commas, but the method presented can easily be extended to other punctuation marks as well. Given the sparsity of n-grams containing commas, the increase of the LM model size is manageable. In the case of the trigram model of (Beeferman et al., 1998), only 14.7% of the trigrams contained commas.

Multiple n-gram models, of different n values, can be combined into polygram models (Schukat-Talamazzini, 1995) by linear interpolation between the models. This allows a polygram model, of higher n value, to exploit the information available in lower-order language models. Equation (1) describes a linear interpolation of trigram, bigram, unigram and zero-gram (1/L) probabilities.

$$P(w|uv) = \rho_0 \frac{1}{L} + \rho_1 P_1(w) + \rho_2 P_2(w|u) + \rho_3 P_3(w|uv) \qquad (1)$$

In this case $\rho_0$, $\rho_1$, $\rho_2$ and $\rho_3$ are the weights associated with the corresponding n-gram probabilities. Other interpolation methods are described in (Schukat-Talamazzini, 1995) and (Schukat-Talamazzini et al., 1997). An example of polygram model usage for sentence segmentation is presented in (Warnke et al., 1997).

A baseline capitalization model based on unigrams is presented in (Lita et al., 2003). In the same paper a more complex model is presented based on a combination of trigrams, bigrams and unigrams as well as a sentence level context.

Stolcke and Shriberg (1997) propose an n-gram based LM for text segmentation. In this case, no actual punctuation is restored, but sentence boundaries without punctuation are detected. This could be seen as a first step towards punctuation restoration. In their LM, the authors included a special token "<s>" indicating the sentence boundary.

In (Niesler and Woodland, 1996) is proposed an n-gram LM based on categories. These models are able to generalize and adapt to unseen word patterns. An example of using such a model is given in (Romero and Sanchez, 2013) where a category-based model is employed to enhance a handwritten text recognition system with emphasis on named entities such as person names, place of residence, geographical origin. As discussed previously, in the "Rules based" section of this survey, named entity detection can play an important role in restoring capitalization.

Agbago et al. (2005) propose enhancing the results of an n-gram model by using a scoring function and incorporating casing probabilities based on word classes for unknown words. These are determined based on word shapes, taking into account the presence of digits or symbols inside a word. Thus, they distinguish between 4 classes: quantity words (starting or ending with numbers), acronyms (containing a sequence of single letters followed by periods), hyphenated words (made up of at least two components joined by a hyphen) and regular uniform words (consisting entirely of alphabetical characters).

Kaufmann (2010) proposed using a trigram model for capitalization of Twitter messages. The author recognizes that original capitalization present in tweets is not a reliable indicator of true capitalization. Users frequently use capitalization when trying to emphasize certain words or ideas. For example, considering an example message (also known as "tweet"): "the movie was very GOOD", we can easily see that the word "GOOD" should not contain any uppercase letter. Most likely, this writing form was used as the user tried to emphasize how much he enjoyed watching the movie. Additionally, the first word should have the first letter uppercase since it is starting a sentence. The proposed LM was built using the Open American National Corpus (OANC) (Ide and Macleod, 2001). In this approach, no actual tweets were used during training.

Nebhi et al. (2015) propose performing truecasing on Twitter texts using a 3-gram language model built using a combination of newswire articles and tweets that were truecased. This approach was taken because not enough training data based solely on tweets was available. Therefore, the addition of newswire articles improved the overall system performance.

Even with today's computing resources, in terms of storage and processing power, models with high n numbers are difficult to handle due to their storage requirements. Therefore, it is common to see relatively low n values (such as 1 through 5). To allow for an easier usage of larger n-gram models, several pruning and compression methods have been proposed, such as (Siivola and Pellom, 2005), (Pauls and Klein, 2011).

A large n-gram language model containing punctuation and casing was gathered by Google, available at the Google Books website[3] and described in (Michel et al., 2010). This model contains n-grams with n ranging from 1 to 5 for multiple languages including English.

Another example of a LM with punctuation included is the Web 1T dataset (Brants and Franz, 2006), also contributed by Google Inc., containing English n-grams (n ranging from 1 to 5) and associated frequency counts calculated over 1 trillion words from web pages collected by Google in January 2006. In addition, the Web 1T 10 European Languages dataset (Brants and Franz, 2009) contains similar data for other languages.

Gravano et al. (2009) studied the impact of increased n-gram order (varying n from 3 to 6) as well as training set size (from 58 million to 55 billion tokens) on both capitalization and punctuation recovery. The focus was on the following punctuation marks: comma, period, question mark and dash. All the other punctuation marks were replaced with one of these marks. After training the LM, a weighted finite state automaton (FSA) is constructed based on the input string considering all possible combinations of capitalization and punctuations. A final decoding step computes the least cost path along the FSA. This is equivalent to the maximum posterior probability sequence of punctuation marks and capitalized words according to the LM. As expected, increasing training set size had a huge impact on performance. Training on the larger corpus produced an increase of 6% in F1 score for capitalization. However, an interesting conclusion was that increasing the n-gram order did not produce such high improvements. Increasing the n-gram order from n=3 to n=6 on both the small and larger corpora yielded and increase in F1 of less than 1% for capitalization. Another conclusion, supporting previous findings, was that low-frequency symbols (like question marks and dashes, that were considered during training) are much harder to model using n-gram based LM.

## 2.3 Capitalization as a discrimination decision

In (Gale et al., 1994) is mentioned that discrimination decisions arise in many natural language processing tasks and the authors make a special mention to the task of deciding whether a word from some teletype text should be capitalized if both cases have been used in the text. Excepting the trivial situation of a word needing capitalization at the beginning of a sentence, it can be easily deduced that a word allowing both lowercase and uppercase usage actually has at least two meanings: one corresponding to the lowercase version and one for the capitalized version. Thus, the matter of deciding which word form to use actually becomes a decision between which word sense to use, hence becoming a word sense disambiguation issue. The authors of (Gale et al., 1994) propose to apply their disambiguation method (Gale et al., 1993) for the purposes of truecasing. This is based on a Bayesian classifier combined with an interpolation procedure, seeking a trade-off between measurement errors and relevance.

Word sense disambiguation has been recognized as a major problem in natural language processing research. This can be encountered in scientific literature as early as Kaplan (1950) and Masterson (1967), which considered that "the basic problem in machine translation is that of multiple meaning". A survey on word sense disambiguation methods is available in (Navigli, 2009).

## 2.4 Hidden event language models

Capitalization can be considered a sequence tagging problem. The sequence is the set of words and the tag set is the capitalization classes as described in the "Introduction" (all letters lowercase, first letter uppercase, all letters uppercase, mixed case). Thus given a word sequence W=$w_0$ $w_1$ $w_2$… $w_n$, the model predicts a sequence of capitalization features C=$c_0$ $c_1$ $c_2$...$c_n$, with $c_i$ from {AL, FU, AU, MC} ("all lower case", "first uppercase", "all uppercase", "mixed case"). Similarly, punctuation restoration can be also considered a sequence tagging problem, predicting inter-word events E=$e_0$ $e_1$ $e_2$ … $e_n$ , where $e_i$ denotes either a punctuation mark or the absence of any marks.

---
[3] http://storage.googleapis.com/books/ngrams/books/datasetsv2.html

Since Maximum Entropy Markov Models (MEMM) were previously used in other sequence tagging problems, such as part-of-speech tagging (Ratnaparkhi, 1996) they can be used for capitalization as well. Considering the set of possible word and tag contexts H and the set of allowable tags T, the probability model can be expressed as:

$$P(h,t) = \pi\mu \prod_{j=1}^{k} \alpha_j^{f_j(h,t)} \quad (2)$$

where h is the sequence history, t is a tag, $\pi$ is a normalization constant, {$\mu, \alpha_1… \alpha_k$} are positive model parameters, $f_1...f_k$ are features with $f_j(h,t) = \{0,1\}$. Under the Maximum Entropy formalism, the goal of the model is to maximize the entropy of a distribution, described by:

$$H(p) = - \sum_{h \epsilon H, t \epsilon T} p(h,t) \log p(h,t) \quad (3)$$

In (Chelba and Acero, 2004) is presented a maximum a-posteriori (MAP) adaptation technique for maximum entropy (MaxEnt) models. This helps reduce the capitalization error of a system trained on a certain type of corpus and used on a different one. The authors give an example of a capitalization system trained on newswire text and used on email and office documents.

Batista et al. (2009) make use of a Maximum Entropy model for capitalization restoration for Spanish and Portuguese languages. The features used are word n-grams of the form: $w_i$, $2w_{i-1}$, $2w_i$, $3w_{i-2}$, $3w_{i-1}$, considering $w_i$ as the current word and $nw_k$ being an n-gram starting at position k. An interesting approach here is the usage of n-grams which are not necessarily centered on the current word. Thus, considering an example sentence "*John went to school in England*", the n-grams used for predicting the case associated with the word "*school*" are: *(school)*, *(to school)*, *(school in)*, *(went to school)*, *(to school in)*. At runtime the authors propose incorporating a lexicon which can provide forced capitalization for previously unseen words during training.

One of the models investigated by Hasan et al., (2014) makes use of a Hidden Markov Model (HMM), similar to the one used previously for disfluencies detection (Stolcke and Shriberg, 1996). This model is only used as a baseline to which more advanced models are compared.

## 2.5 "Boosting" approaches

Boosting (Freund and Schapire, 1995) is a meta-learning approach that aims at combining an ensemble of weak classifiers to form a strong classifier. Considering a black-box learning algorithm, on each iteration t this base learner is used to generate a base classifier $H_t$. The boosting algorithm provides the classifier with the training data as well as a set of weights $W_t$, non-negative, associated with the training example. These weights are used to force the base learner to focus on previously misclassified examples, often considered the "hardest" examples. The output of the final classifier can be considered as:

$$f(x) = \sum_{t=1}^{T} H_t(x) \quad (4)$$

Adaptive Boosting (Adaboost), as described in (Schapire and Singer, 2000) is a greedy search for a linear combination of classifiers by overweighting the examples that are misclassified by each classifier.

Gupta and Bangalore (2002) propose using a boosting approach for building a sentence boundary detector. For this purpose, they use an extension described in (Schapire and Singer, 1999) for multi-class classification problems. The software package used is Boostexter, described thoroughly in (Schapire and Singer, 2000). In this system, the base classifiers output a real number in the interval [-1,+1] whose sign can be interpreted as a prediction regarding the probability of the result belonging to one of the defined classes associated with the result (thus +1 means it belongs with high probability to the respective class).

The real-valued predictions of the final classifier f were converted into probabilities by means of a logistic function:

$$\frac{1}{1 + e^{-f(x)}} \quad (5)$$

This function estimates the probability of x representing a sentence boundary.

As features for the boosting based classifier, Gupta and Bangalore considered a window of three words to the left and three words to the right of the target word. Thus, they used the 3-gram representation on the left side, the right side, the associated part of speech tags on the left and right, as well as the number of common words and part of speech tags considering 1-gram, 2-gram and 3-gram representation on the left and right. This resulted into a total of 18 features, as described in the paper, used for the classifiers.

Furthermore, in (Hasan et al., 2014) is proposed a boosting approach for sentence-end detection, using the ICSIBoost tool[4], which uses an Adaboost algorithm implementation. The textual features used contain n-grams (with n=2,3,4) for the previous words and up to two words following the target word. Additionally, the paper investigates the addition of prosodic features as well as forward and backward distance features.

## 2.6 Conditional Random Fields probabilistic models

Conditional Random Fields (CRF) are probabilistic models used to segment and label sequence data (Lafferty et al., 2001). CRFs present an advantage over maximum entropy Markov models (MEMMs) and other discriminative Markov models based on directed graphical models, which can be biased towards states with few successor states.

Even though CRFs can be used as the final layer of a neural network architecture, this section of the survey will only focus on CRFs used in probabilistic models. Neural networks usage of CRFs is covered in a separated section.

Stanford CoreNLP natural language processing toolkit, as described in (Manning et al., 2014), contains a truecasing module implemented with a discriminative model using a CRF sequence tagger (Finkel et al.,2005). Furthermore, (Wang et al., 2006) presents a bilingual capitalization model for capitalizing machine translation (MT) outputs using CRFs. A characteristic of this model is that it uses both the input sentence and the output sentence of the MT system. In this case the feature functions used include: monolingual language model feature, capitalized translation model feature (given the assumption that the translated word has a high probability of keeping the original word capitalization), capitalization tag translation feature, upper-case translation feature (an all uppercase sequence of words gets translated into a similar all uppercase sequence). Additionally, an initial capitalization feature combined with a punctuation feature template are used to ensure that word is capitalized following a sentence-end punctuation mark.

While considering punctuation restoration for comma, period, question mark and exclamation mark, Lu and Ng (2010) propose using a factorial CRF (F-CRF) model (Sutton et al., 2007). Features are computed using unigrams, bigrams and trigrams. The resulting model is compared against simple n-gram LMs (n=3 and n=5) and linear-chain CRF for both English and Chinese. First of all, these results confirm those of Gravano et al. (2009), which demonstrated that higher order n-gram models do not have a large impact on F1 scores. For both English and Chinese the impact of a larger n-gram model in terms of F1 is less than 1%, in accordance with Gravano's findings. The CRF models are performing better than simple n-gram models, with the F-CRF model having the highest F1 score for both languages.

While continuing the investigation of dynamic CRF models for sentence boundary detection as well as punctuation restoration, Wang et al. (2012) propose using a vocabulary pruning method. This involves selecting the top words appearing in the training data and replacing the others with an unknown token. This reduces the sparsity of the model and makes it more resilient when faced with rare words. Their research suggests an increase in F-measure for punctuation prediction when using around 25% of the words in the initial vocabulary. Ueffing et al. (2013) also tried applying the vocabulary pruning method, but reported "no significant gain". This would seem to indicate that the method's benefits are closely related to the datasets being used for training and testing, especially with regard to the distribution of infrequent words.

---

[4] https://github.com/benob/icsiboost

Hasan et al. (2015) propose training a CRF model on noisy ASR output for sentence end detection. The implementation makes use of the CRF++ toolkit[5] and is explored the impact of noisy textual data for models with both textual only features and additional prosodic features. These models are compared against non-noisy models, trained using official transcripts, and the noisiness seems to improve the model performance on the sentence end detection task. The use of a noisy text-only model on the ASR output text compared to the transcripts model on the same text provides an increase of 2% in terms of F1 evaluation.

## 2.7 Neural network architectures

Architectures based on neural networks for truecasing can be separated based on the prediction given. This can be either at word level, indicating a word class (such as all lowercase, all uppercase, first letter uppercase, mixed case), or at character level, indicating if the individual letter should be lowercase or uppercase. With regard to the punctuation restoration problem, usually punctuation marks are placed between words. However, from a neural network perspective, it is possible to present the network either with complete words or with individual characters. In the first case, the output will be considered as the punctuation mark following an input word, while in the second case the network will predict a punctuation mark when presented with the space character.

Susanto et al. (2016) experimented with two character-level recurrent neural network (RNN) architectures based on either Long Short Term Memory (LSTM) (Hochreiter and Schmidhuber, 1997) or Gated Recurrent Units (GRU) (Cho et al., 2014) cells. They constructed a small model with 2 layers and 300 hidden nodes and a large model with 3 layers and 700 hidden nodes in each layer. Additionally, dropout (Srivastava et al., 2014) is used to reduce the risk of overfitting. Their findings seem to backup the claims of Britz et al. (2017) that LSTMs actually outperforms GRUs. Furthermore, their analysis of the small and large model suggests that a larger RNN model usually improves performance. However, due to possible overfitting, there are situations where a smaller model actually performs better.

In (Ramena et al., 2020) is presented another approach based on a character-level RNN. The network is comprised of 4 layers: a Convolutional Neural Network (CNN) layer that generates character representations, 2 Bidirectional Long Short Term Memory (BiLSTM) layers followed by a final Conditional Random Field (CRF) layer, which decides on the proper case for each of the characters. Furthermore, in the CNN layer, Rectified Linear Unit (ReLU) (Hahnloser et al., 2000; Jarrett et al., 2009) is used for the activation function. Even more, in the CNN and BiLSTM layers an additional dropout is used to improve performance.

Tilk and Alumae (2016) approach the punctuation restoration problem (considering comma, period and question mark) by means of a bidirectional recurrent neural network, implemented using Gated Recurrent Units (GRU) enhanced with an attention mechanism (Bahdanau et al., 2015). As described in the paper, the attention mechanism in this case allows the model to focus on words that may indicate a question but are relatively far apart from the current word. The attention mechanism is incorporated in the model using a late fusion approach, similar to the technique described in (Wang and Cho, 2015). The artificial neural network is comprised of 4 layers: an embedding layer (allowing for both pre-trained vectors or vectors to be trained at runtime), a bidirectional GRU layer (comprised of two sub-layers, one for each direction: forward and reverse), an unidirectional GRU layer with attention and a final SoftMax layer. Training was realized using AdaGrad (Duchi et al., 2011). The results presented in the paper suggest that using a pre-trained vector representation of words may have a beneficial effect on the F1 metric allowing for around 3 percent increase in performance, in the case of using pre-trained GloVe vectors (Pennington et al., 2014) for the English language.

Inspired by Tilk and Alumae's work (2016), Salloum et al. (2017), while working with medical reports, propose implementing a vocabulary reduction pre-processing step. This improves the modeling of rare words and represents a method to deal with out-of-vocabulary words. For this purpose a suffix and prefix list is constructed and each word occurring less than 20 times in the training data is replaced by "pAAAA_n" or "AAAA_ns", where s denotes the suffix, p is the prefix, n is the length of the word without the suffix/prefix rounded to a multiple of 5 letters. In the case of a word containing both a prefix and a suffix from the list, it is actually split into two tokens, one with the prefix and one with the suffix. Finally, if a word is rare but does not have a known prefix or suffix it is represented using the "RARE" token. All these operations take place as a pre-processing step for both training and at runtime. The proposed suffix and prefix list is constructed based on words from a medical terminology as well as some common proper names. This approach seems to have a beneficial effect on the F1 score of each punctuation mark (comma, period, colon). Nevertheless, the results are computed only on the author's medical reports corpus and thus cannot be easily compared with other works.

---

[5] http://crfpp.sourceforge.net/

Klejch et al. (2016) investigate using a sequence-to-sequence mapper (Sutskever et al., 2014) for punctuation restoration. This approach, encountered also in machine translation, is converting a sequence of input words, without punctuation, to a sequence of punctuation marks (period, comma, exclamation mark, question mark, no punctuation).

Gale and Parthasarathy (2017) investigate the usage of character-level neural networks for punctuation prediction and sentence boundary detection. Thus, for each character a punctuation mark (comma, period, question mark) or the lack of punctuation is predicted by the network. The recurrent neural network output is delayed in order to allow the network to capture more context. This is realized by pre-padding the input sequence with n-1 <PAD> tokens and also pre-padding the output sequence with n <PAD> tokens. Thus, a delay of size n is introduced before the first useful tagging decision of the network. Evaluation of the proposed model on two different datasets (which are not publicly available) indicate that the delay character level RNN performance is on par with word-level CRF and RNN models, including word-level delay RNN models.

While investigating sentence segmentation of YouTube automatic subtitles, Song et al. (2019) propose a method based on LSTM networks to add period marks in unsegmented text. For this purpose, the authors used 27826 sentences from online lectures of Stanford University available on YouTube as training data. These were first pre-processed to remove capitalization and any other punctuation except periods. The network architecture used is based on an LSTM layer followed by a SoftMax layer. The input is in the form of Word2Vec (Mikolov et al., 2013) encoded words, while the output is the decision of placing a period after one of the words. The embeddings model used was constructed without capitalization, thus justifying the authors choice for lowercasing the words.

Juin et al. (2017) tackle the problem of punctuation restoration considering a set of 11 output labels: period, comma, question mark, exclamation mark, apostrophe, parentheses (left and right), quotes, semicolon, dash and colon (actually the number of punctuation marks used vary with the datasets employed for experimentation throughout the paper, but the mentioned labels form the maximal set). The neural network model is based on an encoder-decoder architecture, accepting for input the context words together with the associated part-of-speech tags. The actual implementation was realized using the OpenNMT system (Klein et al.,2017), thus handling the problem as a sequence-to-sequence tagging. Given the input word + part of speech tag, the system will output a sequence of words and punctuation marks. The authors employ a pre-trained part-of-speech tagger, which was trained on punctuated text. This is applied on unpunctuated text and unfortunately there is no analysis of the tagging errors due to this high difference between training data vs the actual use case.

Tundik and Szaszak (2018) study the impact of combining a word-based model with a character-based one for restoring punctuation (period, comma, question mark). The first model uses GloVe word embeddings to represent words in a context window and a BILSTM layer followed by a SoftMax layer. The second model makes use of fixed-length sequences of characters, provided as one-hot representations. These are passing through a multi-layer neural network composed of a dense embedding layer, a 1D convolution layer, a max pooling layer and a BILSTM layer followed by a final SoftMax layer. In order to combine the two models, the last two layers (BILSTM and SoftMax) are common. Thus, the word-based model is reduced to the word embeddings and the character-based model only keeps the dense, 1D convolutional and max pooling layers. The authors experimented with individual models and the resulting ensemble model on both English and Hungarian languages. Their findings suggest that performance of the two individual models for English language is within a 3% F1 increase in favor of the word level model, but with better performance of the word-based model on Hungarian language, especially for detecting question marks. Nevertheless, the hybrid model behaves better in both languages, the addition of the character embeddings leading to an increase in F1 and a reduction of the error rate. Finally, the authors also consider the possibility of creating a weighted ensemble of the two models, by using the individual outputs. By choosing appropriate weights, the values can become similar to the hybrid model. However, the hybrid model still outperforms the weighted ensemble.

Yi et al. (2020) is using adversarial learning (Goodfellow et al., 2014) to train a neural network for multitask learning (Caruana, 1997) in order to predict both punctuation and part of speech tagging, using BERT as the base language model. The proposed neural model contains common BERT layers followed by a BiLSTM layer and a CRF layer for each of the classification tasks. Furthermore, the network is completed by an adversarial task discriminator composed of a max-pooling layer, followed by a gradient reversal layer, then a fully connected layer and finally a SoftMax layer. The addition of the adversarial learning mechanism improves the performance of the model, compared with a simple BERT fine tuning approach, with 2% in terms of F1 score for both reference text and ASR output.

Nguyen et al (2019) propose using transformer models with overlapped chunk-merging to restore both capitalization and punctuation in a single pass over the input data, consisting of ASR output. The proposed implementation is split into three components: a chunk splitter which takes a long input sequence and splits it into multiple chunks with overlap, the model itself which predicts simultaneously capitalization and punctuation and a final component which merges the predicted output for the chunks. Chunks are created by using a window of size K tokens sliding with K/2 at each step. The neural network model is implemented based on a transformer architecture, using Tensor2Tensor (Vaswani et al., 2018) and OpenNMT (Klein et al., 2017). The capitalization situations studied include only first letter uppercase and all lowercase, while the punctuation marks considered are comma, period and question mark.

Courtland et al (2020) make use of the RoBERTa model (Liu et al., 2019) as input to a neural network comprised of two additional linear layers. The authors propose aggregating predictions over multiple context windows and show that this approach improves the overall accuracy. This involves predicting punctuation marks (period, comma, question mark) for all the tokens in a window of 50 tokens and then advancing the window with a step of 20 tokens (similar to the method described above of Nguyen et al (2019)). Thus, the tokens will appear in multiple context windows presented to the network and the resulting predictions can be aggregated. This aggregation is performed by a final layer of the network without an external mechanism. The best results are obtained by also fine-tuning the RoBERTa representation on the task data. Nevertheless, the authors also compare pre-trained language models without fine tuning and observe an overall performance with F1 of 83.5 for pre-trained RoBERTa large model comparable to the fine-tuned RoBERTa base model which obtained an F1 score of 83.9. Additionally, very close to these results is a pre-trained XLNet base model which obtained an overall F1 of 82.9.

While working on detecting sentence boundaries in legal texts, Sanchez (2019) observes that legal texts are more challenging than other language data, such as news articles, because most legal texts are structured into sections and subsections, unlike the narrative structure of a news article. Furthermore, legal texts contain additional elements such as header, footer, footnotes or lists and sometimes important sections are interleaved with citations. Even more, specific elements are observed with regard to capitalization restoration, since in legal texts certain named entities (or parts of named entities) are written with all capitals, such as the family name or some institutions. Sanchez (2019) propose using a neural network model based on a BiLSTM hidden layer with a final SoftMax layer to predict for each token if it represents either the beginning of a sentence, an internal token, or the last token of a sentence. The input for the system is comprised of word embeddings (Mikolov, 2013) associated to a three-token window, together with additional binary features indicating the structure of the token (upper or lower case characters, digits, punctuation).

Similar to legal texts, medical texts create problems for general punctuation restoration systems due to their domain specific language. Sunkara et al (2020) propose a combined punctuation and truecasing system for medical texts, especially those obtained by ASR systems from both dictations and conversations. The proposed system has two steps: first punctuation prediction and then the resulting sentence is fed into the capitalization restoration component. The reason for this is the observation that a word following a sentence boundary (indicated by a period) has the first letter uppercase. Thus, the system employs a BERT layer followed by a linear layer with SoftMax activation for punctuation prediction and a final layer linear also with SoftMax activation for capitalization. The final layer makes use of BERT encodings concatenated with the probabilities obtained from the punctuation prediction layer. The BERT model is fine-tuned for the medical domain (Han and Eisenstein, 2019). Afterwards the model is further fine-tuned for task adaptation, by selective masking the input tokens, while ensuring that 50% of the masked tokens are punctuation marks. Finally, in order to account for ASR errors, the model is further augmented by using ASR output aligned with reference text.

## 3 Methods incorporating audio features

Acoustic information involves prosody which comprises features like pause duration between words, pitch-intensity, per-word timing. This is usually used by speakers to impose structure in both spontaneous and read speech (Kompe, 1996). Furthermore, prosodic cues are known to be relevant to discourse structure across languages (Vaissiere, 1983). Even more, in the description of the Verbmobil speech-to-speech translation system, Noth et al. (1999) recognize that "the major role of prosody in human–human communication is segmentation and disambiguation". Additionally, change of speaker can be seen as a helpful information for sentence segmentation since usually one sentence ends and another begins when the speakers change. Based on these observations, when speech data is available, for example in the case of ASR systems, it is desirable to use acoustic information either by itself or together with textual information for punctuation insertion.

Stolcke and Shriberg (1997) proposed an enhanced n-gram model containing information on change of speaker (turn-taking). Thus, they introduced a special token for turn boundaries in both training and testing. They showed that this addition improves over text only n-gram models. Furthermore, in Stolcke et al. (1998) a comparison is performed between a prosodic model, a text-only n-gram model and an n-gram model enhanced with segmentation cues such as turn-taking and long pauses. For both n-gram models it was considered n=4. It is shown that the n-gram model enhanced with segmentation cues derived from speech behaves better for sentence segmentation purposes. The simple text-based n-gram model performs better than the prosodic model on text that was correctly recognized. However, the prosodic model performs better than the text-based n-gram model in case of errors in the ASR output. Nevertheless, in both cases the n-gram model enhanced with speech cues has the highest accuracy. A further improvement is possible by using a model combination of the prosodic model and the n-gram model with segmentation cues via model interpolation technique. However, this only improves the results with less than 1% in terms of accuracy.

In (Warnke et al., 1997) is proposed a combination of a polygram language model and prosodic features used as input for a multilayer perceptron (MLP) classifier for sentence boundary detection. The prosodic features are modeled over a context of six syllables, taking into account duration, pause, F0 and energy.

Chen (1999) propose assigning acoustic baseforms (silence, double silence, disfluency) to punctuation marks into an extended dictionary. Thus, an ASR language model is trained including punctuation marks. At runtime, given a pause or silence detected in the speech, the process will look for the LM score. If this LM score does not indicate a punctuation mark, then it will be interpreted as a silence, otherwise the appropriate punctuation mark will be selected.

Considering the problem of sentence segmentation, Hakkani et al. (1999) explore a large number of prosodic features in both a prosodic-only model and a combined model with word information. Prosodic modelling was realized using CART-style decision trees (Breiman et al., 1984) and the initial features considered are based on the work of Shriberg et al. (1997). By training and evaluating different models with subsets of the initial features, 6 useful features for sentence segmentation were identified: pause duration, F0 (fundamental frequency) differences across the boundary, speaker change, rhyme duration (length of the rhyme, nucleus and coda, of the last syllable preceding the boundary). Out of these features, the most important is indicated to be the pause duration.

While interested in punctuation annotation (comma, period, question mark) of automatically transcribed broadcast news data, Christensen et al. (2001) analyze the impact of different prosody features on finite state and MLP classifiers. Thus, they consider pause durations, phoneme durations (considering the average duration of the phonemes and the average duration of only vowels in the preceding word) and pitch information. Their analysis supports previous research, indicating that pause duration has the biggest impact on the model's accuracy in predicting the correct punctuation mark.

A combination of the previous two approaches is used in Kolář et al. (2004) for building an automatic punctuation annotation system (commas and periods) for Czech language. Thus a textual model is combined with a prosodic model using the following features: F0 derived features (maximum, minimum and mean, first and last value), first and last piecewise linear (PWL) slope, ratio and difference between the last value in the current word and the first value in the following word, ratio and difference between the last PWL slope in the current word and the first PWL slope in the following word, slope of linear regression from all values in the current word), phoneme duration (average duration of vowels, duration of the first and last vowel, duration of the longest and shortest vowel), pause duration, energy features ( maximum, minimum and mean of frame level RMS energy). Two models are explored based on CART decision trees and MLP classification. On the available testing data, it is reported that the CART based model performed better than the MLP model. Unfortunately, the authors did not explore the impact of different prosodic features.

Huang and Zweig (2002) approach the punctuation restoration problem for period, comma and question mark, by using a maximum entropy model with lexical features (a context window of two words before and two words after the current word), the previously identified two punctuation marks, pause duration as a prosody feature. In this experiment, pause duration is measured in 0.01 second intervals and is considered in the 5-word window centered on the current word. They explore both reducing the window size (to only one word before and after the current word) and individual text-only and prosodic-only models. Out of these combinations, the best model is the mixed model (textual and prosody) with two words before and two words after the current word. Nevertheless, the benefit of a larger window translates only in 1% improvement. According to the authors this can be attributed either to the modelling technique used or to the dataset.

Usually, ASR systems are trained using only lowercase words. While trying to handle the capitalization issue, Kim and Woodland (2004) propose training a system using an extended vocabulary, containing both lowercase words and words with the first capital letter. Furthermore, a full stop is modelled based on the silence in speech. This follows their previous work (Kim and Woodland, 2002) where the dictionary was extended using also the all letters uppercase version of words.

Liu et al. (2005) approached the sentence boundary problem using CRFs combining lexical features (such as n-grams and part-of-speech tags) with prosody features, including speaker turn change. The research shows that CRFs have lower error rates compared to HMM and Maxent models trained using similar features. Furthermore, the accuracy of the models can be further increased by using an ensemble mechanism based on a simple voting between the three implemented models.

Batista et al. (2009) used a maximum entropy model for punctuation restoration in Spanish and Portuguese. Their proposed features are $w_i$, $w_{i+1}$, $2w_{i-2}$, $2w_{i-1}$, $2w_i$, $2w_{i+1}$, $3w_{i-2}$, $3w_{i-1}$, $p_i$, $p_{i+1}$, $2p_{i-2}$, $2p_{i-1}$, $2p_i$, $2p_{i+1}$, $3p_{i-2}$, $3p_{i-1}$ (lexical), GenderChgs, SpeakerChgs, and TimeGap (acoustic), where $w_i$ is the current word, $nw_k$ is an n-gram starting at position k, $np_k$ is the part of speech n-gram starting at position k, GenderChgs and SpeakerChgs indicate changes in speaker and gender and TimeGap represents the pause between the current and the next word.

In the context of machine translation of ASR documents, Matusov et al. (2006) propose a sentence segmentation algorithm combining three language models. The first model is based on n-grams and estimates the following probabilities: end of segment boundary, internal segment probability and probability of the first words of the next segment. The second model is based on the pause duration between the assumed last word of a segment and the first word in the next segment. The third model is based on the sentence length probability. Finally, a log-linear combination of the models is constructed using manually tuned scaling factors.

Peitz et al. (2011) further refine on this approach and consider punctuation prediction as a machine translation task. They use a phrase-based MT system (Zens et al., 2008) to align the unpunctuated and punctuated variants of the same text, thus enabling the use of additional features such as phrase translation probabilities. Punctuation is modelled into classes, considering: a class for sentence-end marks (period, question mark, exclamation mark) with sub-classes for period and question mark; a class for commas; and a class for other punctuation marks (quotes, apostrophe, parentheses and semicolon).

Following the evolution of textual only methods for punctuation restoration, as described in other sections of this survey, mixed models using prosody and textual information were adapted to either incorporate the improved language models or to use better classification methods for directly training mixed models. For example, in Kolar and Lamel (2012) is described a system using prosodic features together with text features trained using Adaptive Boosting. Previous findings still hold even with new classification models, since a mixed model performs better than individual text-only or prosodic-only models.

Tilk and Alumae (2015) start by constructing a neural network model using only text features for predicting commas and periods. Then, the last hidden layer of this model is used as input in conjunction with a feature based on the pause duration for training a new model. Both models are using LSTM cells and the usage of the last hidden layer as input for a new model is based on the work of Seide et al. (2011). The two-stage model is completed by a SoftMax layer allowing the prediction of the punctuation mark. Thus, the model learns in the first stage textual features and these are combined in the second stage with the prosodic feature pause duration in order to make better predictions. This approach seems to be beneficial for period restoration, as indicated by the authors. The work was continued in Tilk and Alumae (2016) by replacing the LSTM cells with Bidirectional LSTMs and the addition of an attention mechanism. This led to an F1-score improvement by 2.5% on reference text and 1.8% on ASR output.

Cho et al. (2016) propose using two parallel neural network models, one for text features and one for prosody features, and only combine them in a final stage. The lexical model is trained using only GloVe word embeddings and allows 4 outputs corresponding to a middle word in a context window indicating the punctuation mark (period, comma, question mark) or no punctuation. The acoustic model only considers two output classes: boundary or no boundary. For both models a 3-layer fully connected neural network is used. The final decision for sentence boundary is taken in a 2-stage process. In the first stage both models are used. In this case if an acoustic-based boundary decision is opposed by the lexical model then it will be denied. In the second step, only the lexical model is used together with the output from the first stage. This can lead to further segmentation. Models are evaluated individually and with the decision mechanism after each stage and results prove that the overall model behaves best. The second decision stage adds an increase of 5% in terms of F1 over the first stage decision.

Based on the observation that an ensemble model (Deng and Platt, 2014; Zhao et al., 2015), combining multiple individual models, usually outperforms the individual models, Yi et al. (2017) envisage using such an ensemble model as a teacher for a deep neural network based single model (called a student model). Thus, the models of Tilk and Alumae (2016) and Lample et al. (2016) are combined with a simpler neural network model in order to form the teacher system. Afterwards, the Kullback-Leibler (KL) divergence is used to guide the training of a smaller neural network system. This approach is inspired by the work of Chan et al. (2015) and Chebotar and Waters (2016) which proposed a similar approach for improving speech recognition. Although the resulting punctuation prediction (comma, period, question mark) ensemble model outperforms the student model, the student model is more suitable to deploy in real-life application than the ensemble, due to its simplicity, reflected into smaller size and computing requirements. Furthermore, the student model still outperforms individual models for all the investigated punctuation marks.

Klejch et al. (2017) continue their previous work (Klejch et al., 2016) on sequence-to-sequence models for punctuation prediction by integrating prosody-based features. While most of the previously described models predict a label corresponding to a middle word in a context window, a sequence-to-sequence model uses an encoder-decoder architecture to predict sequences of labels, corresponding to the punctuation marks. These models are usually specific to machine translation applications but in this case the punctuation restoration problem can be considered as a translation from words to punctuation marks. The acoustic model uses a hierarchical encoder. Thus, it transforms frame level acoustic representations into word level acoustic embeddings. The system is trained to predict periods, commas, exclamation marks, question marks and in some experiments also three dots. The inclusion of the acoustic features using this approach leads to an increase in F1 of around 6% for reference text and 3% for ASR output.

Continuing the investigation of encoder-decoder mechanisms for sentence prediction, Yi and Tao (2019) incorporate a self-attention mechanism (Vaswani et al., 2017). Furthermore, in accordance with transfer learning approaches, the authors use pre-trained Word2Vec (Mikolov et al., 2013), GloVe and Speech2Vec (Yu-An and Glass, 2013) representations. The Speech2Vec representation aims to learn a fixed length embedding of an audio segment that captures semantic information of the spoken word directly from audio. The resulting model outperforms previous individual and ensemble-based models (Yi et al, 2017) for restoring periods, commas and question marks.

Zelasko et al. (2018) compare two neural network models based on Bidirectional Long Short-Term Memory (BiLSTM) and Convolutional Neural Networks (CNN) for punctuation prediction (comma, period, question mark). Textual and prosody features were used together in the same model. The corpus used for training and testing consists of telephone conversations, which lead the authors to use a feature for the side pronouncing the words. Additionally, words were encoded using GloVe embeddings and the following audio features were computed: difference between the start of the current word and start of the previous word, and duration of the current word. Thus, contrary to previous research, Zelasko et al. (2018) do not use an explicit pause duration feature. Instead this is encoded in the feature representing the difference between starting time of adjacent words. The evaluation performed concludes that CNN based models yield a better precision while BiLSTM models have better recall. Punctuation predicted by the CNN is more accurate - especially in the case of question marks. Additionally, the authors mention experiments in which they tried to incorporate part of speech tags which didn't lead to any performance improvement.

Szaszak and Tundic (2019) propose combining three neural network models for restoring punctuation (comma, period, question mark, exclamation mark). This is a continuation of their work (Tundik and Szaszak, 2018) where only two models, without prosody, were used. The first model (W) is word-based, using GloVe word embeddings as inputs for a BiLSTM based neural network layer with a final SoftMax layer. The second model (C) is character based (words are represented as a fixed-length sequence of one-hot encoded characters) and is formed of four neural network layers: a character embedding dense layer, a 1D convolutional layer, a BiLSTM layer and a final SoftMax layer. The third model (P) uses prosody features (F0, energy, word duration, pause) as inputs to a BiLSTM layer followed by a SoftMax layer. In order to combine the three models, the final SoftMax layer of each model is removed and the final BILSTM layers are connected into a common LSTM layer which is finally followed by a new SoftMax layer. The authors performed tests using each individual model, combinations of two models and all three of them. For English language, the best results on reference texts are achieved by the combination of all three models (C+W+P), while the best results on ASR output are achieved by the combination W+P. Interestingly, for Hungarian the best results on reference texts are still produced by C+W+P, while on ASR output better results are achieved by C+P. As the authors note, this is due to the agglutination characteristic of the Hungarian language.

Nanchen and Garner (2019) use a Gradient Boosting Machine (GBM) (Friedman, 2000) to improve an ensemble of 4 neural network based algorithms for sentence boundary detection. The first two algorithms make use of textual features, while the next employ pause duration and additional prosody data, obtained using the Kaldi toolkit (Povey et al., 2011) and including probability of voice estimate, pitch and delta-pitch features. By looking at the relative importance metric learned automatically by the GBM algorithm, the authors confirm earlier findings that pause duration is a very important feature for sentence segmentation when audio data is available.

While investigating the performance of punctuation restoration systems when faced with homonymy errors in ASR output, Augustyniak et al. (2020) propose a method for retrofitting pre-trained word embedding representations. This is based on the work of Dingwall and Potts (2018) who suggested updating general pre-trained word embeddings with domain specific information. A comparison between a punctuation restoration model using retrofitted word embeddings fed into a convolutional neural network model against the same model using general word representations shows an improvement of 6% to 9% in terms of F1 score for different punctuation marks.

## 4 Evaluation

Evaluating a capitalization and punctuation restoration system usually involves obtaining a correctly written corpus, which is then pre-processed to remove capitalization and punctuation, then run through the evaluated system and finally the results are compared with the original text. In some cases, when the text corpus is comprised of written transcripts of audio recordings, the text run through the restoration system is the output of an ASR system. From this approach, common evaluation metrics can be computed, such as precision (P), recall (R), F1 measure, Slot Error Rate (SER). However, especially with regard to ASR text, in works such as (Beeferman et al., 1998) is considered that user satisfaction is the most valuable qualitative metric of success for a speech recognition system. Therefore, for the evaluation process, instead of comparing with the original transcript used for performing the recordings, a number of human judges, with different backgrounds, can assess the resulting output.

In an attempt to standardize the evaluation of ASR systems, NIST released a series of rich transcription evaluation data sets[6]. These, or similar attempts at standardizing systems evaluations, could be used also for evaluating capitalization and restoration systems either using ASR output or reference text. However, most systems covered in this survey tend to use dedicated corpora. Thus, it is difficult to compare the performances of different systems in terms of absolute values.

Earlier works, such as (Shriberg et al., 1998) and (Huang and Zweig, 2002) run their evaluation on the Switchboard corpus, available in the Linguistic Data Consortium catalog (LDC97S62). Another set of corpora, popular with earlier works (Hakkani-Tur et al, 1999), (Christensen et al, 2001), (Huang and Zweig, 2002), are Hub4 (LDC98S71) and Hub5 (LDC2002S10).

Recent works are usually relying on datasets from the International Workshop on Spoken Language Translation (IWSLT). These are machine translation datasets that are focused on the automatic transcription and translation of TED and TEDx talks (public speeches covering many different topics)[7]. One dataset that was used in a number of recent works is from the IWSLT 2011 evaluation campaign (Federico et al., 2011). It is comprised of monolingual and parallel corpora of TED talks that are copyright of the TED Conference LLC and distributed under the Creative Commons Attribution-NonCommercial-NoDerivs 3.0 license[8]. The first paper to use this dataset for punctuation prediction evaluation was Peitz et al. (2011). However, as it also happens with later papers, only a fraction of the corpus was used for the punctuation restoration task. In this case, only text documents belonging to the English-to-French translation track.

Peitz et al. (2011) did an analysis of the distribution of punctuation marks in the training and test parts of the used corpus. Their analysis indicates that most of the marks are commas (approximately 50%), followed by periods (39%). Question marks account for roughly 2%, while all the other analyzed punctuation marks (quotes, apostrophe, parentheses and semicolon) represent 9% from the total marks. We can assume that sampling from different parts of the IWSLT 2011 entire corpus will yield similar distributions, thus enabling us to compare various works based on this dataset. Table 1 presents the results reported by different papers using the IWSLT 2011 dataset for evaluation, based on reference text (not ASR output).

**Table 1. Overall evaluation of different works using the IWSLT 2011 dataset**

| Paper | Method | Precision | Recall | F1 |
| --- | --- | --- | --- | --- |
| Peitz et al, 2011 | Statistical machine translation | 80.7 | 64.2 | 71.5 |
| Ueffing et al, 2013 | CRF | 49.8 | 58.0 | 53.5 |
| Tilk and Alumae, 2016 | Bidirectional RNN using GRU | 70.0 | 59.7 | 64.4 |
| Tundik and Szaszak, 2018 | Character + Word based BiLSTM | - | - | 62.9 |

---

[6] https://www.nist.gov/itl/iad/mig/rich-transcription-evaluation

[7] http://www.ted.com

[8] http://creativecommons.org/licenses/by-nc-nd/3.0/

| Yi and Tao, 2019 | Encoder-decoder with self-attention | 76.7 | 69.6 | 72.9 |
| Yi et al., 2020 | Adversarial learning | 80.9 | 75.0 | 77.8 |

In order to better understand the different works included in Table 1, a more detailed evaluation, for each of the punctuation marks common to all papers (period, comma, question mark) is presented in Table 2.

**Table 2. Evaluation on individual punctuation marks (period, comma, question mark) using IWSLT 2011 dataset**

| Paper | Period | | | Comma | | | Question mark | | |
| --- | --- | --- | --- | --- | --- | --- | --- | --- | --- |
| | P | R | F1 | P | R | F1 | P | R | F1 |
| Peitz et al, 2011 | 89.0 | 87.5 | 88.2 | 80.6 | 59.3 | 68.3 | 63.4 | 17.6 | 27.5 |
| Ueffing et al, 2013 | 54.0 | 72.0 | 62.0 | 45.0 | 47.0 | 46.0 | 53.0 | 33.0 | 41.0 |
| Tilk and Alumae, 2016 | 73.3 | 72.5 | 72.9 | 65.5 | 47.1 | 54.8 | 70.7 | 63.0 | 66.7 |
| Tundik and Szaszak, 2018 | 68.4 | 72.6 | 70.4 | 62.4 | 53.7 | 57.7 | 64.1 | 54.3 | 58.8 |
| Yi and Tao, 2019 | 82.5 | 77.4 | 79.9 | 67.4 | 61.1 | 64.1 | 80.1 | 70.2 | 74.8 |
| Yi et al., 2020 | 87.3 | 81.1 | 84.1 | 76.2 | 71.2 | 73.6 | 79.1 | 72.7 | 75.8 |

By looking at the results reported in Tables 1 and 2, it can be seen that the current best overall approach is represented by a neural network implementation using adversarial learning, as described in (Yi et al., 2020). Furthermore, on earlier system (Peitz et al., 2011) even though had a good overall score, it had serious troubles in predicting question marks (an F1 score of 27.5%). Newer systems seem to be able to handle better the reduced amount of question marks present in the training set.

Another observation drawn from the two tables is that comma is the most difficult punctuation mark to predict. Even though, as described earlier, commas account for nearly 50% of the total punctuation marks present in the corpus, neural network approaches seem to have difficulties in properly predicting commas. In neural network based approaches, the F1 score for commas is the lowest between the 3 considered marks. Interestingly, earlier approaches based on statistics (Peitz et al., 2011) and CRF (Ueffing et al, 2013) are worst at predicting question marks, while commas are on the second place. This is in a way expected since question marks represent the least used signs in the training set. However, this also indicates that neural networks are not currently able to grasp the fine details related to comma usage. As mentioned in the introduction, commas may actually belong to multiple categories since they may be able to perform different functions. Unfortunately, there is no analysis of the roles performed by commas in the IWSLT 2011 dataset. Such an analysis may explain better the behavior of neural network approaches.

Even though evaluation metrics such as precision, recall and F1 make sense for the punctuation restoration task, some authors choose to report their results using the slot error rate (SER) metric. In this case we use the formula (6) from Makhoul et al. (1999) to convert the reported SER value into F1.

$$1 - F \cong 0.7 \, SER \quad (6)$$

Punctuation restoration for ASR output text is more challenging. Usually ASR systems introduce errors when compared to reference text. The IWSLT 2011 dataset can be used also for evaluating punctuation restoration on ASR output. Table 3 presents a comparison of recent papers overall F1 score for both reference and ASR output.

**Table 3. Comparison in terms of overall F1 score between punctuation restoration on reference text and ASR output**

| Paper | F1 reference | F1 ASR |
| --- | --- | --- |
| Tilk and Alumae, 2016 | 64.4 | 61.4 |
| Tundik and Szaszak, 2018 | 62.9 | 52.2 |
| Yi and Tao, 2019 | 72.9 | 68.8 |

| Yi et al., 2020 | 77.8 | 73.3 |

From table 3, it results that the system of Yi et al (2020) is the best system also for ASR output, based on the IWSLT 2011 dataset. Furthermore, it can be noticed a roughly 4% decrease in F1 performance for ASR output. A more in-depth analysis of system's performance for ASR output is presented in table 4.

**Table 4. Evaluation on individual punctuation marks (period, comma, question mark) on ASR output, using the IWSLT 2011 dataset**

| Paper | Period | | | Comma | | | Question mark | | |
|---|---|---|---|---|---|---|---|---|---|
| | P | R | F1 | P | R | F1 | P | R | F1 |
| Tilk and Alumae, 2016 | 70.7 | 72.0 | 71.4 | 59.6 | 42.9 | 49.9 | 60.7 | 48.6 | 54.0 |
| Tundik and Szaszak, 2018 | 63.5 | 67.1 | 65.3 | 50.1 | 48.2 | 49.1 | 60.0 | 42.9 | 50.0 |
| Yi and Tao, 2019 | 75.5 | 75.8 | 75.6 | 64.0 | 59.6 | 61.7 | 72.6 | 65.9 | 69.1 |
| Yi et al., 2020 | 80.0 | 79.1 | 79.5 | 72.4 | 69.3 | 70.8 | 68.4 | 66.0 | 67.2 |

When comparing table 4 with table 2, the same drop in performance can be seen for each individual punctuation mark. One slight difference can be observed with regard to comma vs question mark performance. If in table 2 the comma was the hardest punctuation mark to predict for all systems, on ASR output in table 4, the system of Yi et al (2020) seems to be actually getting better at predicting commas and having more difficulties at predicting question marks.

In both reference text and ASR output, the performance associated with restoring periods is always the best by a large margin of more than 10% in terms of F1 score compared to commas. In Figure 1 is presented the evolution in terms of F1 score difference between periods and commas on reference text. Adversarial learning employed by Yi et al (2020) is improving the performance of comma prediction and reduces the F1 difference to period prediction from 20% (in the case of Peitz et al (2011)) to 10.5%.

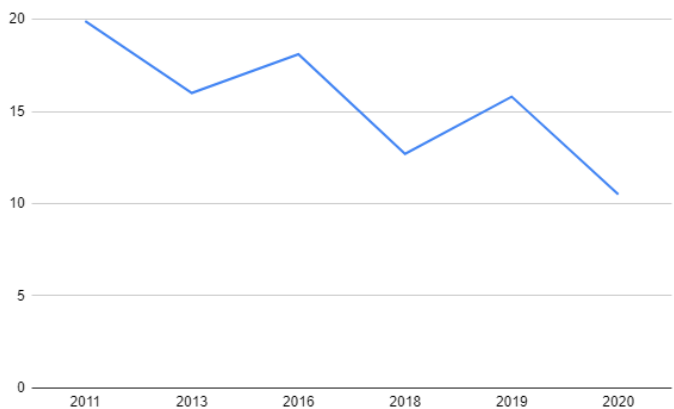

**Figure 1: Difference in restoration performance in terms of F1 score between periods and commas**

Further considering the problem of comma prediction, it seems that all the systems have higher precision and lower recall values (for both reference and ASR output). This indicates a higher number of false negatives (lack of comma prediction). A similar phenomenon takes place in the case of question mark prediction.
Similar to punctuation restoration, works on capitalization made use of different datasets, which makes a proper comparison of earlier systems to more recent ones to be quite difficult to realize. Earlier datasets used for capitalization evaluation include NIST 1998 Hub-4 (Pallett et al, 1998), WSJ PTB3 (Marcus et al, 1999), CSR HUB4 (MacIntyre, 1998). More recent works, such as Susanto et al (2016) and Ramena et al (2020) make use of a Wikipedia based corpus (Coster and Kauchak, 2011). Furthermore, the authors of Ramena et al (2020) also evaluated on this Wikipedia corpus two earlier systems, that of Lita et al (2003) and Stanford CoreNLP (Manning et al, 2014). The F1 scores are presented in Table 5.

**Table 5. Comparison in terms of F1 score between capitalization restoration systems**

| System | F1 |
|---|---|
| Lita et al, 2003 | 86.8 |
| Stanford CoreNLP (Manning et al, 2014) | 90.9 |
| Susanto et al, 2016 | 93.2 |
| Ramena et al, 2020 | 94.0 |

The two best performing systems from Table 5, (Susanto et al, 2016) and (Ramena et al, 2020) make use of character-level recurrent neural network architectures. In the first paper is employed a model based on 3 LSTM layers, while in the second paper a more complex network is built using a convolutional layer, two BiLSTM layers and a final CRF layer.The added neural network complexity only accounts for 0.8% increase in the F1 score. The reported precision (94.9) and recall (93.1) are quite similar and there is no clear indication of where the network fails to correctly identify the correct case.

Apart from general narrative datasets, specialized datasets were used. Thus, Sanchez (2019) used the dataset initially introduced by Savelka (2017) to implement and assess boundary detection systems for legal text. The author reports an F1 score of 89 using a neural network approach.

Sunkara et al (2020) make use of two medical domain datasets: a dictation corpus consisting of 3.7M words and a conversational corpus containing 51M words. Their transformer-based model with BERT embeddings manage to obtain an F1 score of 79 for period and 69 for comma on the conversational corpus while on the dictation corpus the scores are 85 for period and 72 for comma. A similar difference exists also for the capitalization task, where the model achieves an F1 of 83 on the conversational corpus and 87 on the dictation corpus for identifying words having their first letter uppercase. When comparing these results with those presented in the tables above, it becomes clear that domain specific data poses significant challenges for both punctuation restoration and truecasing architectures.

## 5 Conclusions

This survey presented the various problems and techniques associated with the tasks of restoring proper punctuation and capitalization. These are important first steps when annotating text from sources missing or having incomplete punctuation or capitalization, such as automatic speech recognition systems, certain micro-blogging activities, short text messages, some raw OCR output. We began by describing the tasks and the motivation behind them, then described methods using only lexical features, followed by methods incorporating audio features. Afterwards, we presented evaluation considerations with regard to datasets used in scientific papers as well as a comparison of current systems on common datasets.

Already in 1997, Stolcke and Shriberg recognized that if "part-of-speech information is to be used for segmentation, an automatic tagging step is required. This presents somewhat of a chicken-and-egg problem, in that taggers typically rely on segmentations." A possible solution to this problem would be to train both segmentation and POS tagging as a single process. A step in this direction was taken by Yi et al. (2020) who used adversarial learning for training a neural network to predict both punctuation and POS tagging, using BERT as the base language model.

Long range dependencies are often required for punctuation restoration (Lu and Ng, 2010). This is a clear disadvantage with models concerned only with local context, such as n-gram models or even neural networks using small context windows. In this case recurrent neural network models (such as those based on BiLSTM cells) seem to have a clear advantage, being able to make use of longer contexts.

With the evolution of classification algorithms and language modelling methods, better models were created for punctuation restoration. As text-only models evolved so did mixed models, combining audio information with word data. However, one thing remained constant: mixed models are performing better than individual ones (text-only or prosody-only). This is confirmed in numerous papers cited throughout this survey where such individual models were compared with mixed models.

Current state-of-the-art models for both capitalization and punctuation restoration make use of neural network architectures. Considering the papers of Yi et al (2020), for punctuation restoration, and Ramena et al (2020), for truecasing, both make use of BiLSTM layers with a final CRF layer. Huang et al (2015) investigated the usage of an architecture based on BiLSTM+CRF for different natural language processing tasks. Their findings suggest that this architecture is able to achieve state-of-the-art results on part of speech tagging, chunking and named entity recognition. Furthermore, the models are less dependent on word embeddings compared to other neural network architectures.

A special case for punctuation restoration is represented by the task of detecting sentence boundaries. This seems a somewhat simpler situation in which the system does not have to predict the exact punctuation at the end of sentence but instead it has to recognized that a sentence ends and another one begins. Even though in some domains this is considered a solved problem, there are still domains like legal text which present unique challenges (Sanchez, 2019). This observation can then be extended to the more difficult tasks of complete punctuation restoration and capitalization. In the case of legal texts, also capitalization is sometimes a problem for regular punctuation restoration systems, since certain named entities (such as family names or certain institutions) are written with all uppercase letters (Sanchez, 2019; Savelka et al., 2017). This creates new research challenges for creating restoration systems for legal texts and also to identify other domains where general purpose systems may not offer sufficiently good results. In this context, the work of Sunkara et al (2020) investigates medical domain texts and argues for the necessity of domain adaptation techniques that would allow improvement of punctuation restoration and truecasing systems.

ASR systems often produce recognition errors. In most of the papers presented in section 3 of this survey, the impact of ASR errors can be seen on punctuation restoration systems in the form of algorithm performance differences between execution against ASR output and reference text. As indicated by Augustyniak et al. (2020) while investigating homonymic errors one possibility to improve punctuation restoration systems is in the form of better word embeddings representations. However, since these are not the only errors occurring in ASR systems, future research should focus also on different types of errors.Even though the English language has 14 punctuation marks (period, question mark, exclamation, comma, semicolon, colon, dash, hyphen, brackets, braces, parentheses, apostrophe, quotation marks, ellipsis) most of the works on punctuation restoration are limited to only three of them (period, question mark and comma), with a few exceptions such as Juin et al (2017) working with 8 punctuation marks, Peitz et al (2011) working with 3 marks but also considering a special "others" class of punctuation, Gravano et al (2009) considering the dash, Salloum et al (2017) considering the colon, Klejch et al (2017) considering exclamation mark and ellipsis. This seems to indicate that more research is required for proper restoration of other punctuation marks apart from period, question mark and comma.